\documentclass{article} % For LaTeX2e
\usepackage[dvips]{graphicx}
\usepackage{nips14submit_e,times}
\usepackage{hyperref}
\usepackage{mdframed}
\usepackage{rotating}

\usepackage[outdir=./]{epstopdf}

\usepackage{graphics,graphicx}
\usepackage{amsmath,amssymb,amsfonts,comment}

\newcommand{\p}{\mathrm{P}}
\newcommand{\word}[1]{\texttt{#1}}

\urldef{\googlegroup}\url{https://groups.google.com/forum/#!forum/word2vec-toolkit}
\urldef{\blogpost}\url{https://blog.lateral.io/2015/06/the-unknown-perils-of-mining-wikipedia/}

\title{Controlled Experiments for Word Embeddings}

 \author{
 	Benjamin Wilson\\
	Lateral GmbH\\
	\texttt{benjamin@lateral.io}
	\And
	Adriaan M.\,J.\, Schakel\\
	NNLP\\
	\texttt{adriaan.schakel@gmail.com}
 }

\date{\today}
\nipsfinalcopy % Uncomment for camera-ready version

\begin{document}

\maketitle

\begin{abstract}
	An experimental approach to studying the properties of word
        embeddings is proposed.  Controlled experiments, achieved
        through modifications of the training corpus, permit the
        demonstration of direct relations between word properties and
        word vector direction and length.  The approach is demonstrated
        using the word2vec CBOW model with experiments that
        independently vary word frequency and word co-occurrence noise.
        The experiments reveal that word vector length depends more or
        less linearly on both word frequency and the level of noise in
        the co-occurrence distribution of the word.  The coefficients of
        linearity depend upon the word.  The special point in feature
        space, defined by the (artificial) word with pure noise in its
        co-occurrence distribution, is found to be small but non-zero.
\end{abstract} 

\begin{section}{Introduction}
Word embeddings, or distributed representations of words, have been the
subject of much recent research in the natural language processing and
machine learning communities, demonstrating state-of-the-art performance
on word similarity and word analogy tasks, amongst others.  Word
embeddings represent words from the vocabulary as dense, real-valued
vectors.  Instead of one-hot vectors that merely indicate the location
of a word in the vocabulary, dense vectors of dimension much smaller
than the vocabulary size are constructed such that they carry syntactic
and semantic information.  Irrespective of the technique chosen, word
embeddings are typically derived from word co-occurrences.  More
specifically, in a machine-learning setting, word embeddings are
typically trained by scanning a short window over all the text in a
corpus.  This process can be seen as sampling word co-occurrence
distributions, where it is recalled that the co-occurrence distribution
of a target word $w$ denotes the conditional probability $\p(w'|w)$ that
a word $w'$ occurs in its context, i.e., given that $w$ occurred.  Most
applications of word embeddings explore not the word vectors themselves,
but relations between them to solve, for example, similarity and word
relation tasks \cite{dontcountpredict}.  For these tasks, it
was found that using normalised word vectors improves performance.  Word
vector length is therefore typically ignored.

In a previous paper \cite{schakel-wilson}, we proposed the use of word
vector length as measure of word significance.  Using a domain-specific
corpus of scientific abstracts, we observed that words that appear only
in similar contexts tend to have longer vectors than words of the same
frequency that appear in a wide variety of contexts.  For a given
frequency band, we found meaningless function words clearly separated
from proper nouns, each of which typically carries the meaning of a
distinctive context in this corpus.  In other words, the longer its
vector, the more significant a word is.  We also observed that word
significance is not the only factor determining the length of a word
vector, also the frequency with which a word occurs plays an important
role.

In this paper, we wish to study in detail to what extent these two
factors determine word vectors.  For a given corpus, both term
frequency and co-occurrence are, of course, fixed and it is not obvious
how to unravel these dependencies in an unambiguous, objective manner.
In particular, it is difficult to establish the distinctiveness of the
contexts in which a word is used.  To overcome these problems, we
propose to modify the training corpus in a controlled fashion.  To
this end, we insert new tokens into the corpus with varying frequencies and
varying levels of noise in their co-occurrence distributions.  By modeling the
frequency and co-occurrence distributions of these tokens, or 
\textit{pseudowords}\footnote{We refer to these tokens as
pseudowords, since their properties are modeled upon words in the
lexicon and because our corpus modification approach is reminiscent of the
pseudoword approach for generating labeled data for word sense disambiguation
tasks in \cite{gale1992work}.}, on existing words in the corpus, we are able to
study their effect on word vectors independently of one another.  We can thus
study a family of pseudowords that all appear in the same context, but with
different frequencies, or study a family of pseudowords that all have the same
frequency, but appear in a different number of contexts.  Starting from the
limited number of contexts in which a word appears in the original corpus, we
can increase this number by interspersing the word in arbitrary contexts at
random.  The word thus looses its significance in a controlled way.  Although
we present our approach using the word2vec CBOW model, these and related
experiments could equally well be carried out for other word embedding methods
such as the word2vec skip-gram model \cite{DistRepns,EfficientEstimation},
GloVe \cite{pennington2014glove}, and SENNA \cite{collobert-2011}.

We show that the length of the word vectors generated by the CBOW model
depends more or less linearly on both word frequency and level of noise
in the co-occurrence distribution of the word.  In both cases, the
coefficient of linearity depends upon the word.  If the co-occurrence
distribution is fixed, then word vector length increases with word
frequency.  If, on the other hand, word frequency is held constant, then
word vector length decreases as the level of noise in the co-occurrence
distribution of the word is increased.  In addition, we show that the direction
of a word vector varies smoothly with word frequency and the level of
co-occurrence noise.  When noise is added to the
co-occurrence distribution of a word, the corresponding vector smoothly
interpolates between the original word vector and a small vector
perpendicular to it that represents a word with pure noise in its
co-occurrence distribution.  Surprisingly, the special point in feature
space, obtained by interspersing a pseudoword uniformly at random
throughout the corpus with a frequency sufficiently large to sample all
contexts, is non-zero.

This paper is structured as follows.  Section~\ref{related-work} draws
connections to related work, while Section~\ref{corpus-and-model} describes
the corpus and the CBOW model used in our experiments.
Section~\ref{WFVE} describes a controlled experiment for varying word
frequency while holding the co-occurrence distribution fixed.
Section~\ref{CNVE}, in a complementary fashion, describes a controlled
experiment for varying the level of noise in the co-occurrence
distribution of a word while holding the word frequency fixed.  The
final section, Section~\ref{future-directions}, considers further questions
and possible future directions.
\end{section}

\begin{section}{Related work}\label{related-work}
Our experimental finding that word vector length decreases with co-occurrence
noise is related to earlier work by Vecchi, Baroni, and Zamparelli
\cite{vecchi-baroni-zamparelli2011}, where a relation between vector length and
the ``semantic deviance'' of an adjective-noun composite was studied
empirically.  In that paper, which is also based on word co-occurrence
statistics, the authors study adjective-noun composites.  They built a
vocabulary from the 8k most frequent nouns and 4k most frequent adjectives in a
large general language corpus and added 22k adjective-noun composites.  For
each item in the vocabulary, they recorded the co-occurrences with the top 10k
most frequent content words (nouns, adjectives or verbs), and constructed word
embeddings via singular value decomposition of the co-occurrence matrix
\cite{landauer-dumais1997}.  The authors considered several models for
constructing vectors of unattested adjective-noun composites, the two simplest
being adding and component-wise multiplying the adjective and noun vectors.
They hypothesized that the length of the vectors thus constructed can be used
to distinguish acceptable and semantically deviant adjective-noun composites.
Using a few hundred adjective-noun composites selected by humans for
evaluation, they found that deviant composites have a shorter vector than
acceptable ones, in accordance with their expectation.  In contrast to their
work, our approach does not require human annotation.

Recent theoretical work \cite{Arora2015} has approached the problem of
explaining the so-called ``compositionality'' property exhibited by some word
embeddings.  In that work, unnormalised vectors are used in their model of the
word relation task.  It is hoped that experimental approaches such as those
described here might enable theoretical investigations to describe the role of
the word vector length in the word relation tasks.  \end{section}

\begin{section}{Corpus and model}\label{corpus-and-model} Our training data is
	built from the Wikipedia data dump from October 2013.  To remove the
	bulk of robot-generated pages from the training data, only pages with
	at least 20 monthly page views are retained.\footnote{For further
	justification and to obtain the dataset, see\\ \blogpost} Stubs and
	disambiguation pages are also removed, leaving 463 thousand pages with
	a total of 482 million words.  Punctuation marks and numbers were
	removed from the pages and all words were lower-cased.  Word
	frequencies are summarised in Table~\ref{word-occurrences-table}.  This
	base corpus is then modified as described in Sections~\ref{WFVE} and
	\ref{CNVE}.  For recognisability, the pseudowords inserted into the
	corpus are upper-cased.

\begin{subsection}{Word2vec}\label{word2vec} Word2vec, a feed-forward neural
	network with a single hidden layer, learns word vectors from word
	co-occurrences in an unsupervised manner.  Word2vec comes in two
	versions.  In the continuous bag-of-words (CBOW) model, the words
	appearing around a target word serve as input.  That input is projected
	linearly onto the hidden layer and the network then attempts to predict
	the target word on output.  Training is achieved through
	back-propagation.  The word vectors are encoded in the weights of the
	first synaptic layer, ``syn0''.  The weights of the second synaptic
	layer (``syn1neg'', in the case of negative sampling) are typically
	discarded.  In the other model, called skip-gram, target and context
	words swap places, so that the target word now serves as input, while
	the network attempts to predict the context words on output.

For simplicity only the word2vec CBOW word embedding with a single set of
hyperparameters is considered.  Specifically, a CBOW model with a hidden layer
of size 100 is trained using negative sampling with 5 negative samples, a
window size of $10$, a minimum frequency of $128$, and $10$ passes through the
corpus.  Sub-sampling was not used so that the influence of word frequency
could be more clearly discerned.  Similar experimental results were obtained
using hierarchical softmax, but these are omitted for succinctness.  The
relatively high low-frequency cut-off is chosen to ensure that word vectors, in
all but degenerate cases, receive a sufficient number of gradient updates to be
meaningful.  This frequency cut-off results in a vocabulary of $81117$ words
(only unigrams were considered).

The most recent revision of word2vec was used.\footnote{SVN revision 42, see
\url{http://word2vec.googlecode.com/svn/trunk/}} The source code for performing
the experiments is made available on
GitHub.\footnote{\url{https://github.com/benjaminwilson/word2vec-norm-experiments}}

\begin{table}
	\begin{tabular}{c | r | l}
frequency band & \# words & example words  \\
\hline
$2^{0} - 2^{1}$ & 979187 & \word{isa220, zhangzhongzhu, yewell, gxgr} \\
$2^{1} - 2^{2}$ & 416549 & \word{wz132, prabhanjna, fesh, rudick} \\
$2^{2} - 2^{3}$ & 220573 & \word{gustafsdotter, summerfields, autodata, nagassarium} \\
$2^{3} - 2^{4}$ & 134870 & \word{futu, abertillery, shikaras, yuppy} \\
$2^{4} - 2^{5}$ & 90755 & \word{chuva, waffling, wws, andujar} \\
$2^{5} - 2^{6}$ & 62581 & \word{nagini, sultanah, charrette, wndy} \\
$2^{6} - 2^{7}$ & 41359 & \word{shew, düül, kidjo, strangeways} \\
$2^{7} - 2^{8}$ & 27480 & \word{smartly, sydow, beek, falsify} \\
$2^{8} - 2^{9}$ & 17817 & \word{legionaries, möbius, mannerism, cathars} \\
$2^{9} - 2^{10}$ & 12291 & \word{bedtime, disabling, jockeys, brougham} \\
$2^{10} - 2^{11}$ & 8215 & \word{frederic, monmouth, constituting, grabbing} \\
$2^{11} - 2^{12}$ & 5509 & \word{questionable, bosnian, pigment, coaster} \\
$2^{12} - 2^{13}$ & 3809 & \word{dismissal, torpedo, coordinates, stays} \\
$2^{13} - 2^{14}$ & 2474 & \word{liberty, hebrew, survival, muscles} \\
$2^{14} - 2^{15}$ & 1579 & \word{destruction, trophy, patrick, seats} \\
$2^{15} - 2^{16}$ & 943 & \word{draft, wood, ireland, reason} \\
$2^{16} - 2^{17}$ & 495 & \word{brought, move, sometimes, away} \\
$2^{17} - 2^{18}$ & 221 & \word{february, children, college, see} \\
$2^{18} - 2^{19}$ & 83 & \word{music, life, following, game} \\
$2^{19} - 2^{20}$ & 29 & \word{during, time, other, she} \\
$2^{20} - 2^{21}$ & 17 & \word{has, its, but, an} \\
$2^{21} - 2^{22}$ & 10 & \word{by, on, it, his} \\
$2^{22} - 2^{23}$ & 4 & \word{was, is, as, for} \\
$2^{23} - 2^{24}$ & 3 & \word{in, and, to} \\
$2^{24} - 2^{25}$ & 1 & \word{of} \\
$2^{25} - 2^{26}$ & 1 & \word{the} \\
\end{tabular}

	\caption{ Number of words, by frequency band, as observed in the
          unmodified corpus.  }
	\label{word-occurrences-table}
\end{table}
\end{subsection}

\begin{subsection}{Replacement procedure}\label{replacement-procedure}
In the experiments detailed below, we modify the corpus in a controlled
manner by introducing pseudowords into the corpus via a replacement
procedure.  For the frequency experiment, the procedure is as follows.
Consider a word, say \word{cat}.  For each occurrence of this word, a
sample $i$, $1 \leqslant i \leqslant n$ is drawn from a truncated
geometric distribution, and that occurrence of the word \word{cat} is
replaced with the pseudoword \word{CAT\_i}.  In this way, the word \word{cat}
is replaced throughout the corpus by a family of pseudowords with varying
frequencies but approximately the same co-occurrence distribution as
\word{cat}.  That is, all these pseudowords are used in roughly the same
contexts as the original word.

The geometric distribution is truncated to limit the number of pseudowords
inserted into the corpus.  For any choice $0 < p < 1$ and maximum value
$n > 0$, the truncated geometric distribution is given by the
probability density function
\begin{equation} 
\label{distro}
P_{p, n} (i) = \frac{p^{i-1}  (1-p)}{1 - p^n}, \qquad 1
\leqslant i \leqslant n.
\end{equation} 
The factor in the denominator, which tends to unity in the limit $n \to
\infty$, assures proper normalisation.  We have chosen this distribution
because the probabilities decay exponentially base $p$ as a function of
$i$.  Of course, other distributions might equally well have been chosen
for the experiments.

For the noise experiment, we take instead of a geometric distribution,
the distribution
\begin{equation}
\label{distro2} 
P_n (i) = \frac{2 (n-i)}{n(n-1)} , \qquad 1 \leqslant i \leqslant n.
\end{equation} 
We have chosen this distribution for the noise experiment, because it
leads to evenly spaced proportions of co-occurrence noise that cover the entire interval
$[0,1]$.
\end{subsection}
\end{section}

\begin{section}{Varying word frequency}\label{WFVE}
In this first experiment, we investigate the effect of word frequency on
the word embedding.  Using the replacement procedure, we introduce a
small number of families of pseudowords into the corpus.  The pseudowords in each
family vary in frequency but, replacing a single word, all share a
common co-occurrence distribution.  This allows us to study the role of
word frequency in isolation, everything else being kept equal.  We
consider two types of pseudowords.

\begin{subsection}{Pseudowords derived from existing words}\label{WFVEexisting}
We choose uniformly at random a small number of words from the
unmodified vocabulary for our experiment.  In order that the inserted
pseudowords do not have too low a frequency, only words which occur at least
10 thousand times are chosen.  We also include the high-frequency
stopword \word{the} for comparison.  Table~\ref{word-frequency-counts}
lists the words chosen for this experiment along with their frequencies.

The replacement procedure of Section~\ref{replacement-procedure} is then
performed for each of these words, using a geometric decay rate of $p =
\tfrac{1}{2}$, and maximum value $n=20$, so that the 1st pseudoword is
inserted with a probability of about $0.5$, the 2nd with a probability of
about $0.25$, and so on.  This value of $p$ is one of a range of values
that ensure that, for each word, multiple pseudowords will be inserted with a
frequency sufficient to survive the low-frequency cut-off of
128.  A maximum value $n=20$ suffices for this choice of $p$, since
$2^{20 + \log_2{128}}$ exceeds the maximum frequency of any word in
the corpus.  Figure~\ref{fig:word-frequency-experiment-text-cat}
illustrates the effect of these modifications on a sample text, with a
family of pseudowords \word{CAT\_i}, derived from the word \word{cat}.
Notice that all occurrences of the word \word{cat} have been replaced
with the pseudowords \word{CAT\_i}.

\begin{table}
	\begin{center}
\begin{tabular}{l | r}
word & frequency \\
\hline
\word{lawsuit} & 11565 \\
\word{mercury} & 13059 \\
\word{protestant} & 13404 \\
\word{hidden} & 15736 \\
\word{squad} & 24872 \\
\word{kong} & 32674 \\
\word{awarded} & 55528 \\
\word{response} & 69511 \\
\word{the} & 38012326 \\
\end{tabular}
\end{center}

	\caption{Words chosen for the word frequency experiment, along with their frequency in the unmodified corpus. }
	\label{word-frequency-counts}
\end{table}

\begin{figure}
	\begin{mdframed}
	\input{word-frequency-experiment-text-cat.tex}
	\end{mdframed}
	\caption{Example sentences modified in the word frequency
          experiment as per Section~\ref{WFVEexisting}, where the word
          \word{cat} is replaced with pseudowords \word{CAT\_i} using the
          truncated geometric distribution~(\ref{distro}) with
          $p=\tfrac{1}{2}$ and $n=20$.}
	\label{fig:word-frequency-experiment-text-cat}
\end{figure}

\end{subsection}

\begin{subsection}{Pseudowords derived from an artificial, meaningless word}
\label{WFVEmeaningless}
Whereas the pseudowords introduced above all replace an existing word that
carries a meaning, we now include for comparison a high-frequency,
meaningless word.  We choose to introduce an artificial, entirely
meaningless word \word{VOID} into the corpus, rather than choose an
existing (stop)word whose meaninglessness is only supposed.  To achieve
this, we intersperse the word uniformly at random throughout the corpus
so that its relative frequency is $0.005$.  The co-occurrence
distribution of \word{VOID} thus coincides with the unconditional word
distribution.  The replacement procedure is then performed for this
word, using the same values for $p$ and $n$ as above.
Figure~\ref{fig:word-frequency-experiment-text-void} shows the effect of
these modifications on a sample text, where a higher relative frequency
of $0.05$ is used instead for illustrative purposes.
%Although the probability of inserting the pseudoword
%\word{VOID\_1} is twice that of inserting the pseudoword \word{VOID\_2} and four
%times that of inserting the pseudoword \word{VOID\_3}, these three pseudowords all
%appear once in this small piece of text.

\begin{figure}
	\begin{mdframed}
	\input{word-frequency-experiment-text-void.tex}
	\end{mdframed}
	\caption{The same example sentences as in
          Figure~\ref{fig:word-frequency-experiment-text-cat} where
          instead of the word \word{cat} now the meaningless word
          \word{VOID} is replaced with pseudowords \word{VOID\_i}.  For
          illustrative purposes, the meaningless word \word{VOID} was
          here interspersed with a relative frequency of
          $0.05$. }
	\label{fig:word-frequency-experiment-text-void}
\end{figure}
\end{subsection}

\subsection{Experimental results}\label{WFVE-results}
We next present the results of the word frequency experiment. We
consider the effect of word frequency on the direction and on the length
of word vectors separately.

\subsubsection{Word frequency and vector direction}\label{WFVE-direction}
Figure~\ref{word-frequency-experiment-heatmap} shows the cosine
similarity of pairs of vectors representing some of the pseudowords used in
this experiment.  Recall that the cosine similarity measures the extent
to which two vectors have the same direction, taking a maximum value of
$1$ and a minimum value of $-1$.  The number of different pseudowords
associated with an experiment word is the number of times that its
frequency can be halved and remain above the low-frequency cut-off of
$128$.

Consider first the vectors for the pseudowords associated to the word
\word{the}.  Notice that the cosine similarity of the vectors for
\word{THE\_1} and \word{THE\_i} decreases monotonically with $i$, while the
cosine similarity of the vectors for \word{THE\_i} and \word{THE\_18}
increases monotonically with $i$.  Indeed the direction of
the vector \word{THE\_i} changes systematically, interpolating between the
directions of the vectors of the highest-frequency pseudoword
\word{THE\_1} and the lowest-frequency pseudoword \word{THE\_18}.  The same
trend is apparent (though over shorter frequency ranges) for all the families of pseudowords
other than that for \word{VOID}.

Consider now the vectors for pseudowords derived from the meaningless word
\word{VOID}.  The vectors for \word{VOID\_7}, \dots, \word{VOID\_13}
are approximately orthogonal to one another, just as would be expected from
randomly drawn vectors in a high dimensional space.
As the pseudoword \word{VOID} occurs by
construction in every context, a much higher number of samples is
required to capture its co-occurrence distribution, and
thereby to learn its vector (the same is true, but to a lesser
extent, for the stopword \word{the}).
We conclude that the vectors corresponding to the lower frequency
pseudowords \word{VOID\_7}, \dots, \word{VOID\_13} have not been trained on a
sufficient number of samples to establish their proper direction.
These vectors are excluded from further analysis.
The vectors for \word{VOID\_1}, \dots, \word{VOID\_6}, on the other hand,
exhibit the smooth change in vector direction with word frequency described in
the previous paragraph.

In recent work on the evaluation of word embeddings, Schnabel et
al.~\cite{schnabelemnlp2015} trained logistic regression models to
predict whether a word was rare or frequent given only the direction of
its word vector.  For various word embedding methods, the prediction
accuracy was measured as a function of the threshold for word rarity.
It was found in the case of word2vec CBOW that word vector direction
could be used to distinguish very rare words from all other words.
Figure~\ref{word-frequency-experiment-heatmap} is consistent with this finding,
as it is apparent that word vector direction does change gradually with
frequency.  Schnabel et al.\ claim further that word vector direction must
encode word frequency directly, and not indirectly via semantic information.
Figure~\ref{word-frequency-experiment-heatmap}, considered for any particular
experiment word in isolation (e.g. \word{SQUAD}), demonstrates that the variance of word
vector direction with word frequency is indeed independent
of co-occurrence (semantic) information, and thereby provides further evidence
for this claim. 

\subsubsection{Word frequency and vector length}
We next consider the effect of frequency on word vector length.
Throughout, we measure vector length using the Euclidean norm.
Figure~\ref{fig:word-frequency-experiment-graph} shows this relation for
individual words, both for the word vectors, represented by the weights
of the first synaptic layer, syn0, in the word2vec neural network, and
for the vectors represented by the weights of the second synaptic layer,
syn1neg.  We include the latter, which are typically ignored, for
completeness.  Each line corresponds to a single word, and the points on
each line indicate the frequency and vector length of the pseudowords derived
from that word.  For example, the six points on the line corresponding
to the word \word{protestant} are labeled, from right to left, by the
pseudowords \word{PROTESTANT\_1}, \word{PROTESTANT\_2}, \dots,
\word{PROTESTANT\_6}.  Again, the number of points on the line is
determined by the frequency of the original word.  For example, the
frequency of the word \word{protestant} can be halved at most $6$ times
so that the frequency of the last pseudoword is still above the low-frequency
cut-off.  Because all the points on a line share the same co-occurrence
distribution, the left panel in
Figure~\ref{fig:word-frequency-experiment-graph} demonstrates
conclusively that length does indeed depend on frequency directly.
Moreover, this relation is seen to be approximately linear for each word
considered.  Notice also that the relative positions of the lengths of
the word vectors associated with the experiment words are roughly
independent of the frequency band, i.e., the plotted lines rarely cross.

Observe that the lengths of the vectors representing the meaningless pseudowords
\word{VOID\_i} are approximately constant (about $2.5$).  Since we
already found the direction to be also constant, it is sensible to speak
of the word vector of \word{VOID} irrespective of its frequency.  In
particular, the vector of the pseudoword \word{VOID\_1} may be taken as an approximation.

\begin{figure}
	\includegraphics[scale=0.5]{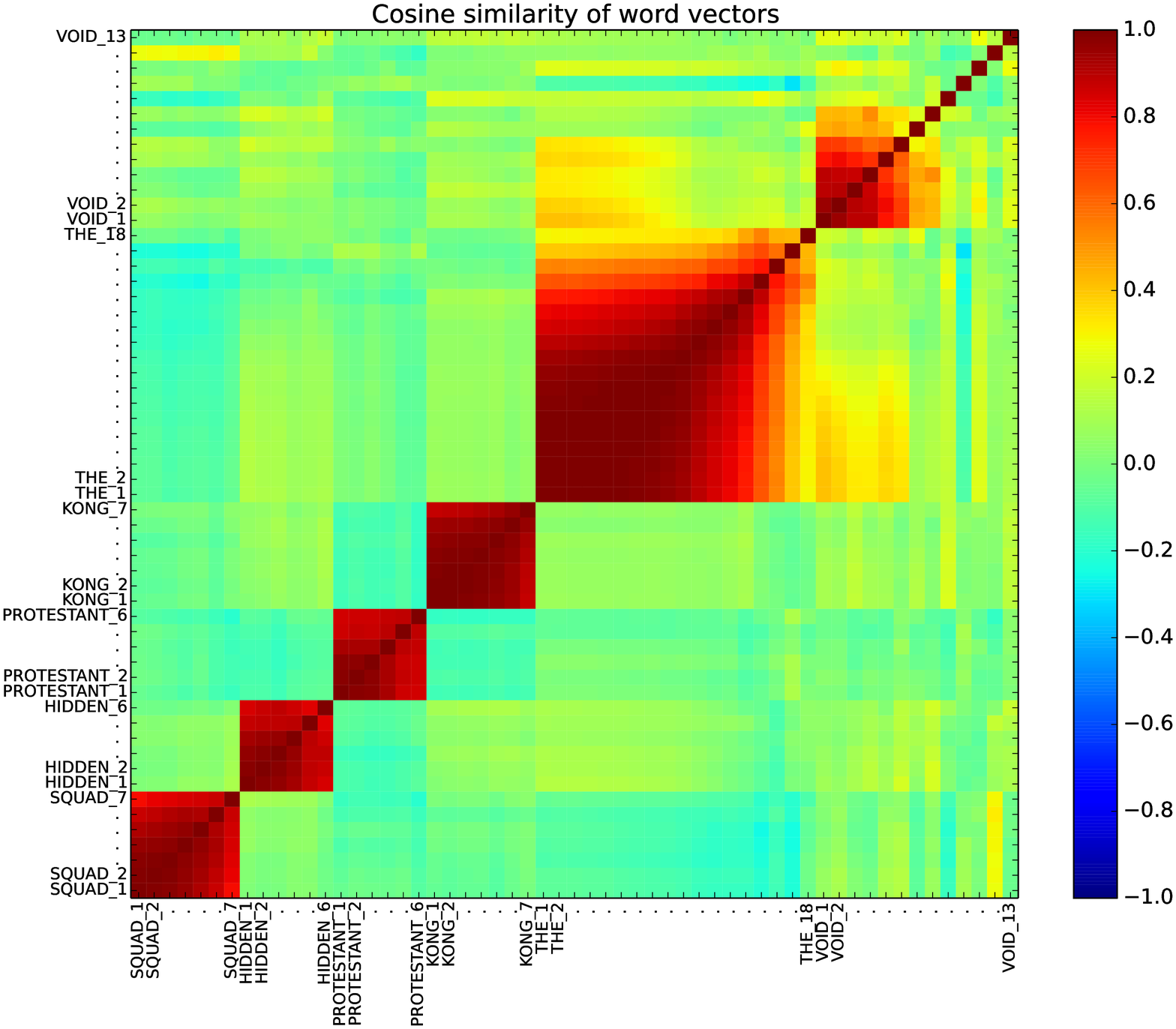}
	\caption{ Heatmap of the cosine similarity of the vectors
          representing some of the pseudowords used in the word frequency
          experiment.  The words other than \word{the} and \word{VOID}
          were chosen randomly.  }
	\label{word-frequency-experiment-heatmap}
\end{figure}

\begin{sidewaysfigure*}
	\centering{\includegraphics[scale=0.6]{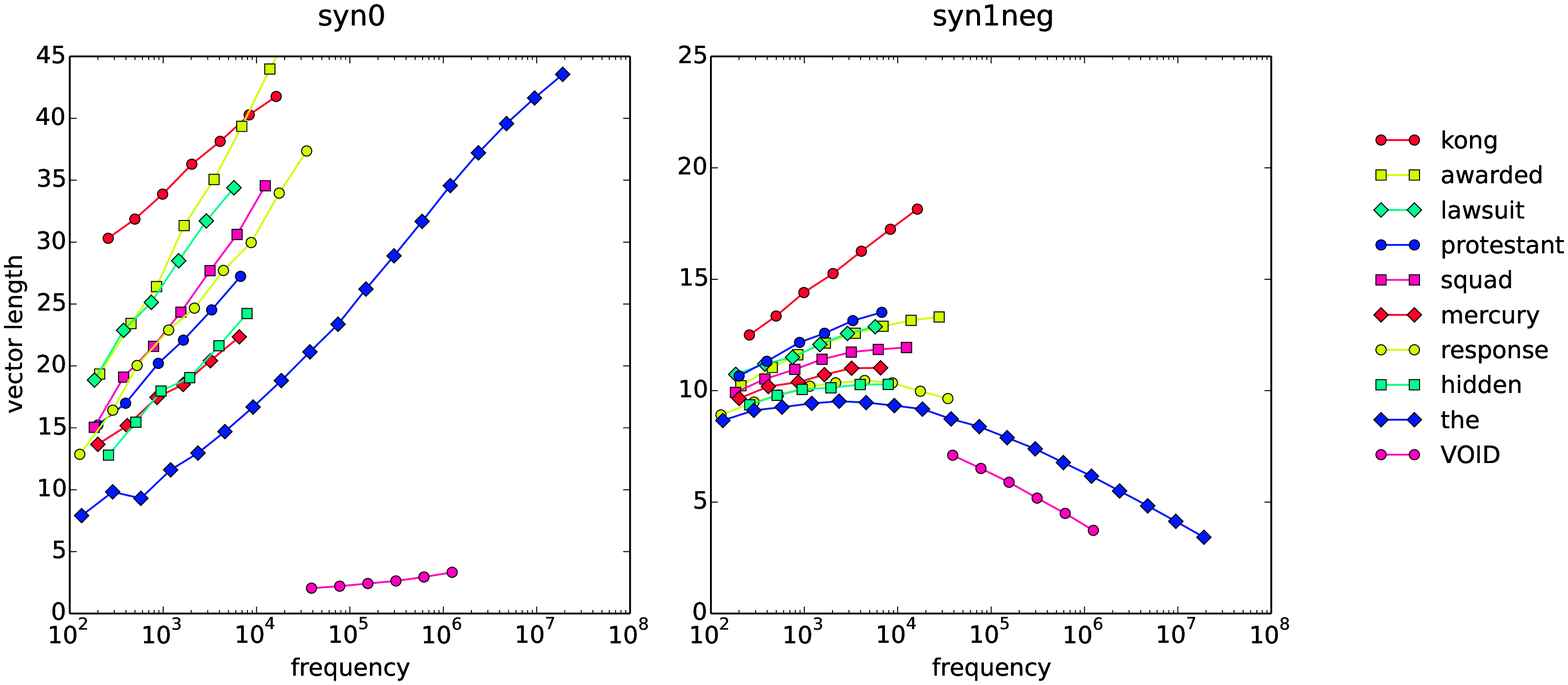}}
	\caption{ Vector length vs.\ frequency for pseudowords derived from a
          few words chosen at random.  For each word, pseudowords of varying
          frequency but with the co-occurrence distribution of that word
          were inserted into the corpus, as described in
          Section~\ref{WFVE}.  The vectors are obtained from the first
          synaptic layer, syn0, of the word2vec neural network.  The
          vectors obtained from the second layer, syn1neg, are included
          for completeness.  Legend entries are ordered by vector length
          of the left-most data point in the syn0 plot, descending.}
	\label{fig:word-frequency-experiment-graph}
\end{sidewaysfigure*}
\end{section}

\begin{section}{Varying co-occurrence noise}\label{CNVE}
This second experiment is complementary to the first.  Whereas in the
first experiment we studied the effect of word frequency on word vectors
for fixed co-occurrence, we here study the effect of co-occurrence noise
when the frequency is fixed.  As before, we do so in a controlled
manner.
  
\subsection{Generating noise}
We take the noise distribution to be the (observed) unconditional word
distribution.  Noise can then be added to the co-occurrence distribution
of a word by simply interspersing occurrences of that word uniformly at
random throughout the corpus.  A word that is consistently used in a
distinctive context in the unmodified corpus thus appears in the
modified corpus also in completely unrelated contexts.  As in
Section~\ref{WFVE}, we choose a small number of words from the
unmodified corpus for this experiment.
Table~\ref{cooccurrence-noise-words} lists the words chosen, along with
their frequencies in the corpus.

For each of these words, the replacement procedure of
Section~\ref{replacement-procedure} is performed using the
distribution~(\ref{distro2}) with $n=7$.  For every replacement pseudoword
(e.g. \word{CAT\_i}), additional occurrences of this pseudoword are
interspersed uniformly at random throughout the corpus, such that the
final frequency of the replacement pseudoword is $2/n$ times that of the
original word \word{cat}.  For example, if the original word \word{cat}
occurred $1000$ times, then after the replacement procedure,
\word{CAT\_2} occurs approximately $238$ times, so a further
(approximately) $2/7 \times 1000 - 238 \approx 48$ random occurrences of
\word{CAT\_2} are interspersed throughout the corpus.  In this way, the
word \word{cat} is removed from the corpus and replaced with a family of
pseudowords \word{CAT\_i}, $1 \leqslant i \leqslant 7$.  These pseudowords all
have the same frequency, but their co-occurrence distributions, while
based on that of \word{cat}, have an increasing amount of noise.
Specifically, the proportion of noise for the $i$th pseudoword is
$$ 1 - \frac{n}{2} P_n(i) = \frac{i-1}{n-1} , \;\; \textrm{ or } \;\; 0,
\frac{1}{n-1}, \frac{2}{n-1}, \ldots, 1 \;\; \textrm{ for } \;\; i =
1,2, \ldots, n , 
$$ 
which is evenly distributed.  The first pseudoword contains no noise at all,
while the last pseudoword stands for pure noise.  The particular choice of
$n$ assures a reasonable coverage of the interval $[0,1]$.  Other
parameter values (or indeed other distributions) could, of course, have
been used equally well.

Figure~\ref{fig:co-occurrence-noise-experiment-text} illustrates the
effect of this modification in the case where the only word chosen is
\word{cat}.  The original text in this case concerned both cats and
dogs.  Notice that the word \word{cat} has been replaced entirely in the
cats section by \word{CAT\_i} and, moreover, that these same pseudowords
appear also in the dogs section. These occurrences (and additionally,
with probability, some occurrences from the cats section) constitute
noise.

\begin{table}
	\begin{center}
\begin{tabular}{l | r}
word & frequency \\
\hline
\word{dying} & 10693 \\
\word{bridges} & 12193 \\
\word{appointment} & 12546 \\
\word{aids} & 13487 \\
\word{boss} & 14105 \\
\word{removal} & 15505 \\
\word{jobs} & 21065 \\
\word{community} & 115802 \\
\end{tabular}
\end{center}

	\label{fig:co-occurrence-noise-counts}
	\caption{Words chosen for the co-occurrence noise experiment,
          along with the word frequencies in the unmodified corpus. }
	\label{cooccurrence-noise-words}
\end{table}

\begin{figure}
	\begin{mdframed}
	\input{cooccurrence-noise-experiment.tex}
	\end{mdframed}
	\caption{Example sentences modified for the co-occurrence noise
          experiment, where the word \word{cat} was chosen for
          replacement.  The pseudowords were generated using the
          distribution~(\ref{distro2}) with $n=7$.}
	\label{fig:co-occurrence-noise-experiment-text}
\end{figure}

\begin{subsection}{Experimental results}
Figure~\ref{fig:co-occurrence-noise-heatmap} shows the cosine similarity
of pairs of vectors representing some of the pseudowords used in this
experiment.  Remember that the first pseudoword~($i=1$) in a family is
without noise in its co-occurrence distribution, while the last one
($i=n$, with $n=7$) stands for pure noise and has therefore no relation
anymore with the word it derives from.  The figure demonstrates that the
vectors within a family only moderately deviate from the original
direction defined by the first pseudoword ($i=1$) when noise is added to the
co-occurrence distribution.  For $1<i<7$, the deviation typically
increases with the proportion of noise.  The vector of the last pseudoword
($i=n$), associated with pure noise, is seen within each of the families
to point in a completely different direction, more or less perpendicular
to the original one.  To understand this interpolating behavior, recall
from Section~\ref{WFVE-results} that the vector for the entirely
meaningless word \word{VOID} is small but non-zero.  Since the noise
distribution coincides with the co-occurrence distribution of
\word{VOID}, the vectors for the experiment words must tend to the word
vector for \word{VOID} as the proportion of noise in their co-occurrence
distributions approaches $1$.
This convergence to a common point is only indistinctly apparent in
Figure~\ref{fig:co-occurrence-noise-heatmap}, as the frequency of the experiment
pseudowords is insufficient to sample the full variety of the contexts of
\word{VOID}, i.e., all contexts (see Section~\ref{WFVE-direction}).

The left panel in Figure~\ref{fig:co-occurrence-noise-graph} reveals
that vector length varies more or less linearly with the proportion of
noise in the co-occurrence distribution of the word.  This figure
motivates an interpretation of vector length, within a sufficiently
narrow frequency band, as a measure of the absence of co-occurrence
noise, or put differently, of the extent to which a word carries the
meaning of a distinctive context.

\begin{figure}
	\includegraphics[scale=0.5]{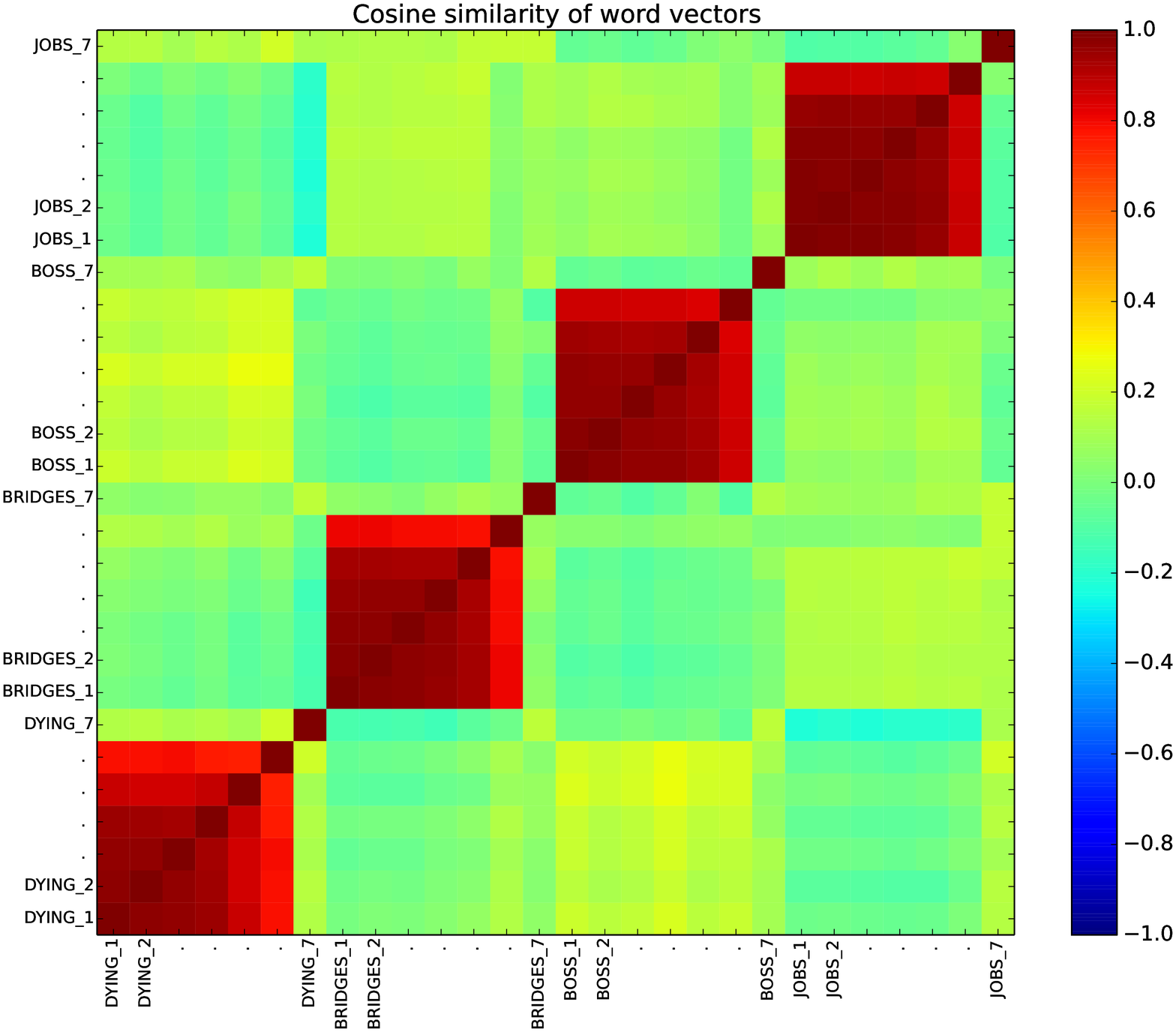}
	\caption{ Heatmap of the cosine similarity of the vectors
          representing some of the pseudowords used in the co-occurrence
          noise experiment (the words were chosen at random).  The
          largely red blocks demonstrate that for $i<7$ the direction of
          the vectors only moderately changes when noise is added to the
          co-occurrence distribution.  The vector of the pseudowords
          associated with pure noise ($i=7$) is seen to be almost
          perpendicular to the word vectors they derive from. }
	\label{fig:co-occurrence-noise-heatmap}
\end{figure}

\begin{sidewaysfigure*}
	\includegraphics[scale=0.6]{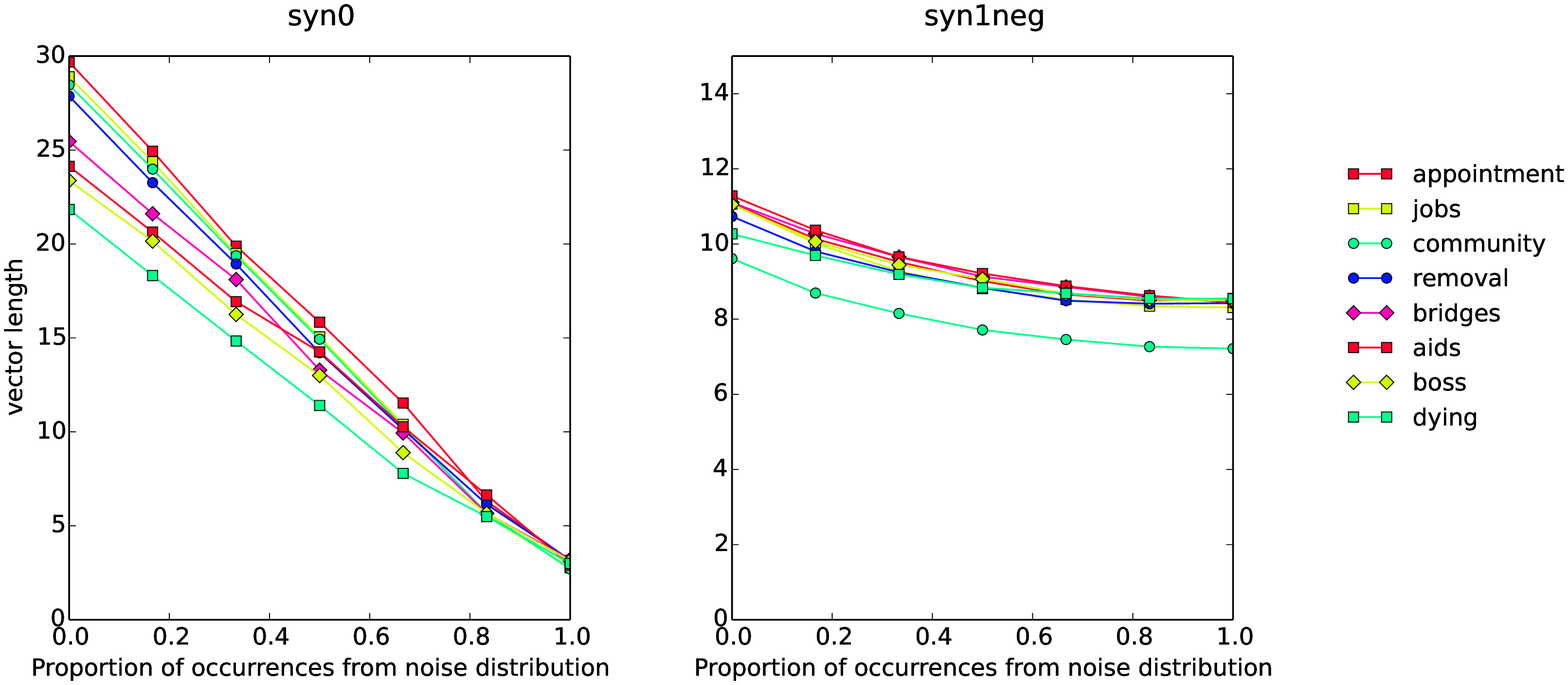}
	\caption{ Vector length vs.\ proportion of occurrences from the noise distribution
          for words chosen for this experiment.  For each word, pseudowords
          of equal frequency but with increasing proportion of
          co-occurrence noise were inserted into the corpus, as
          described in Section~\ref{CNVE}.  The word vectors are
          obtained from the first synaptic layer, syn0.  The second
          layer, syn1neg, is included for completeness.  Legend entries
          are ordered by vector length of the left-most data point in
          the syn0 plot, descending.}
	\label{fig:co-occurrence-noise-graph}
\end{sidewaysfigure*}

\end{subsection}

\end{section}

\begin{section}{Discussion}\label{future-directions}
Our principle contribution has been to demonstrate that controlled
experiments can be used to gain insight into a word embedding.  These
experiments can be carried out for any word embedding (or indeed
language model), for they are achieved via modification of the training
corpus only.  They do not require knowledge of the model implementation.
It would naturally be of interest to perform these experiments for other
word embeddings other than word2vec CBOW, such as skipgrams and GloVe,
as well as for different hyperparameters settings.

More elaborate experiments could be carried out.  For instance, by
introducing pseudowords into the corpus that mix, with varying proportions,
the co-occurrence distributions of two words, the path between the word
vectors in the feature space could be studied.  The co-occurrence noise
experiment described here would be a special case of such an
experiment where one of the two words was \word{VOID}.

Questions pertaining to word2vec in particular arise naturally from the
results of the experiments.
Figures~\ref{fig:word-frequency-experiment-graph} and
\ref{fig:co-occurrence-noise-graph}, for example, demonstrate that the
word vectors obtained from the first synaptic layer, syn0, have very
different properties from those that could be obtained from the second
layer, syn1neg.  These differences warrant further investigation.

The co-occurrence distribution of \word{VOID} is the unconditional
frequency distribution, and in this sense pure background noise.  Thus
the word vector of \word{VOID} is a special point in the feature space.
Figure~\ref{fig:word-frequency-experiment-graph} shows that this point
is not at the origin of the feature space, i.e., is not the zero vector.
The origin, however, is implicitly the point of reference in word2vec
word similarity tasks.  This raises the question of whether improved
performance on similarity tasks could be achieved by transforming the
feature space or modifying the model such that the representation of
pure noise, i.e., the vector for \word{VOID}, is at the origin of the
transformed feature space.
\end{section}

\begin{section}{Acknowledgments}
The authors thank Tobias Schnabel for helpful discussions.
\end{section}

\clearpage
\footnotesize
\bibliography{main}
\bibliographystyle{plain}
\end{document}